# Interplay: Dispersed Activation in Neural Networks

*Richard L. Churchill*

## Introduction

The modeling of neural networks as simple *n*-by-*n* matrices in a Hebbian-learning scheme [1], presents certain problems with regard to the temporal activation sequence of those neurons, both with regard to fragment based recall and ordering due to spatial relations. The Kak model [2] for full recall of memories based upon partial cues (B-matrices) partially addresses these issues, but there are two questions related to this model that deserve some attention.

The first question is regarding the spatial relationship themselves [3]. Is a 2D model of distance a sufficient representation for some or all neural network simulations, and if only for some, which? It seems apparent that some forms of neural activities are reasonably well represented by a 2D model of distance. For example, the sense of touch in a confined area of skin starts with activation of sensory neurons in an essentially 2D region. The sense of smell originates in a confined area of the nose called the olfactory epithalamium that is a surface that is at least in small regions reasonably 2D. Vision, or at least the more precise portion of the field of vision is also a surface, but the stimulus provided by light makes the representation of the stimulus over a confined area of the retina effectively 2D. However, once we start dealing with "deeper" brain function, the questions are more complex. While the cerebral cortex is a surface over much of the brain, it is both highly convoluted and thick, making a 3D model of distance appear more appropriate. When dealing with complex interactions between memory and multiple sensory stimuli, the need to model distance as a 3D property seems even more appropriate.

While such considerations may appear inconsequential, the temporal effects of distance on the interplay of neural stimuli upon memory and recall should not be discounted. For the purposes of what follows, we will assume that a 3D modeling of distance is appropriate, as we are dealing with questions of interplay between "remote" initial stimuli, and the interactions that ensue, in considering conscious recall of memories.

The second question is regarding the interplay between independent partial-stimuli originating from neurons or neuron clusters that are "remote" from each other. This question is the primary focus of this paper. We will examine only simple cases with stimulus events arising at only two points or regions in the network as a starting point for possible subsequent explorations.

Connectivity between neurons is accomplished via synapses at the terminal ends of dendrites [3],[4]. A single neuron may have tens of thousands of dendrites, and the routing of these dendrites among the neurons and other dendrites in brain matter is of concern in 3D modeling of neural networks. "Shortest distance" routing cannot be assumed, but cannot be easily modeled. Therefore, we ignore it with the assumption that the effects of such routing problems are reasonably uniform among sets of connected neurons, and directly related to the distances between the neuron cell nuclei.

For purposes of simplicity we also confine the following discussion to small neural networks, containing only 20 neurons, rather than tackle the problems at full scale. Such is possible, but involves considerations beyond the scope of an initial discussion of a concept, and more resources than available at the time this exploration was conducted.

While we could make the simplifying assumption that our neural network is composed of neurons uniformly distributed through a parallelepiped, the resulting distances would include significant numbers of "ties" at progressively increasing distances from a given neuron. We do not see this as at all appropriate, since temporal issues are involved. However, again for purposes of simplifying an initial examination, we do not address finely divided time intervals, only layering of intervals in a coarse sense. Thus, the step between activation of a neuron and its stimulation of a subsequent neuron is dealt with only in terms of the distance-sequence from the originating neuron. More accurate, of course, would be activation sequenced by nearest-neighbor from the active set to the inactive set, corrected for dendritic activations related to neuron stimulation and other propagation delays. That problem is left to others, or to later examinations.

Please take specific note of the fact that in the following we assume that it makes sense to speak of stimulation of a neuron in terms of both +1 and -1 stimulations. This explicit assumption is based on the belief that such quantum-like effects do exist in the nervous system, and are distinct from non-stimulation, which we would treat as a 0. While this does not account for all possible quantum-like interactions and events in such a system (including true quantum processes [5],[6], as well others that are spatially and structurally defined [7]), it does at least allow us to raise some questions that may be considered in future work.

## Generating a 3D Network

While the generation of a virtual 3D network is simple, it is appropriate to demonstrate that it was done in a reasonable manner. We randomly generated a set of three 20-element vectors, each representing the location of one neuron corresponding to the index for the entry in one axis of 3D space. For easy of publishing, we used integer coordinates in the range [0, 9] for all three axes. The coordinate vectors were as follows for one of the trials.

X = [4, 9, 1, 8, 6, 3, 1, 4, 4, 1, 5, 2, 3, 5, 2, 2, 6, 2, 8, 9]
Y = [7, 3, 5, 1, 9, 8, 8, 2, 5, 0, 4, 3, 1, 1, 4, 0, 5, 4, 6, 6]
Z = [6, 0, 0, 3, 5, 6, 4, 8, 7, 9, 5, 3, 1, 6, 7, 4, 0, 2, 1, 2]

We then constructed from this a right upper triangular matrix, with entries containing the distance from a neuron indicated by row number to one corresponding to the column number [8],[9]. Adding this matrix to its transpose then provides the full matrix of distances, and we can freely look up the distance from any neuron to any other neuron. The validity of the resulting distance vectors in terms of no co-locations of neurons is easily checked by looking for zero distances not on the main diagonal of the matrix, and cases where co-locations occur can be discarded.

As we are dealing with only two-point stimulations within the network, it is a simple matter to either randomly or arbitrarily select two starting points/neurons within the network and construct the required permutation matrices to remap the matrix to use either starting point as the basis for stimulation sequences.

Two neurons were selected as initial points of stimulation. The method of selection was by the minimum and maximum x-axis coordinates in their first appearances in the X vector. For the sample trial presented here, these turned out to be neurons 2 and 3. These arbitrary selections helped assure us of some separation between these neurons, and thus a convergence spread over a larger set of neurons.

With the two initial neurons selected, the rows in the distance matrix corresponding to these neurons were sorted for increasing, preserving the original indices to create a permutation vector pair, which are as follows for the presented example.

pix1 = [2 19 17 20 4 13 11 18 14 12 3 5 16 1 9 8 6 15 7 10]

pix2 = [3 18 12 13 7 17 11 16 1 6 15 19 9 5 2 14 20 4 8 10]

From these vectors, it is easy to see that the first overlaps between the stimulation processes will occur at neurons 18, 12, 17, 13 and 11. Our initial working hypothesis was that the converging stimulus processes will merge smoothly if the stimuli are consistent with the same memories, and that if single-point stimuli do not so perform then slightly larger stimulus events (involving one or two neurons closest to the initially selected neurons) will. The specific example presented here is in part chosen for the second neuron (18) in the second sequence being located at position 8 of the first sequence.

Using the permutation matrices for the selected neurons, two B-Matrices were generated corresponding to the required stimulation sequences.

The "memories" generated for this trial are the following three, with the corresponding permuted memories consistent with the selected initial neurons and stimulation sequences.

rMems' = [-1  1  1  1  1 -1 -1 -1  1 -1 -1 -1 -1  1 -1  1 -1  1  1  1;
           1  1  1 -1 -1  1  1  1 -1 -1  1 -1 -1 -1  1 -1  1 -1  1 -1;
          -1  1 -1  1 -1  1  1 -1 -1  1 -1  1 -1  1 -1 -1  1  1  1  1]

r1Mems' = [1  1 -1  1  1 -1 -1  1  1 -1  1  1  1 -1  1 -1 -1 -1 -1 -1;
           1  1  1 -1 -1 -1  1 -1 -1 -1  1 -1 -1  1 -1  1  1  1  1 -1;
           1  1  1  1  1 -1 -1  1  1  1 -1 -1 -1 -1 -1  1 -1  1  1]

r2Mems' = [1  1 -1 -1 -1 -1 -1  1 -1 -1 -1  1  1  1  1  1  1  1 -1 -1;
           1 -1 -1 -1  1  1  1 -1  1  1  1  1 -1 -1  1 -1 -1 -1  1 -1;
          -1  1  1 -1  1  1 -1 -1 -1  1 -1  1 -1 -1  1  1  1  1 -1  1]

From these memories a neural network was constructed by the normal methods. Two additional networks were constructed from this these memories by applying the generated permutation sequences to the initial network, and these were then reduced to B-matrices appropriate to the two initial neurons.

The original rMems generated matrix was tested to verify that memories were in fact retrievable using the Kak B-matrix method, which proved to be the case. The same was applied to the permuted memory generated B-matrices, and again memories were retrievable. In the case of the r1Mems set, only one of the three memories was retrieved by this method, all three starting with a 1 in the initial neuron position. The memory retrieved was the third, with the iterative process actually closing on the solution for this memory well ahead of the clamped positions.

For r2Mems, the retrieved memory with an initial stimulus vector [1 0 0 0 0 0 0 0 0 0 0 0 0 0 0 0 0 0 0 0] was the second. Using an initial stimulus vector of [-1 0 0 0 0 0 0 0 0 0 0 0 0 0 0 0 0 0 0 0] (which corresponds to the third memory in the r2Mems permutations), the method failed to close on the third memory, and instead produced the pseudo-memory [-1 1 1 1 1 1 -1 -1 -1 1 -1 1 -1 -1 1 1 1 1 -1 1]. Notice that this pseudo-memory is actually only one value off from the third memory, and that in the fourth neuron in the sequence. Such results occurred in several trials, but were not common. As such, the phenomenon deserves study, particularly as it appears to relate directly to flawed memories that are very nearly accurate.

In the present case, it was decided that this phenomenon was worth pursuing with regard to the interplay of dispersed, sparse stimulation.  As the r1Mems trial accurately and rapidly closed on the third memory, and the r2Mems set produced a near miss, it was considered worth investigating whether the interplay might correct the flawed memory, or produce the same or a different flaw.  Given the ordering of the neuron stimulation sequence in the r2Mems set, the fourth position was expected to lead to clamping the sixth position in memory 3 of the r1Mems sequence incorrectly, with an anticipated cascade of flaws from there.

## Interplay

In order to study the interplay between relatively remote stimulus events we continue by applying the B-matrix clamping technique simultaneously to both of the stimulation sequences (r1Mems and r2Mems).

Neuron 18 is clamped in the r2Mems sequence.  At this point, it must be observed that since this clamping occurs well in advance of the clamping process for the r1Mems sequence, a decision must be made whether to handle the cascade from that stimulation event outside the r1Mems and r2Mems sequences.  It must be expected that neuron 18 will relay to its nearest neighbor after some latency interval.  The answer to that question is not simple.

Generating the stimulation sequence for neuron 18, we get the following ordering: 18, 12, 3, 13, 11, 16, 7, 17, 15, 1, 9, 6, 14, 19, 8, 4, 5, 20, 2 and 10.  Note also that neuron 12 is the nearest neighbor of 18, followed by 3.  This last is not a surprise, since 18 is neuron 3's nearest neighbor.  But also, 12 is the second neuron after 18 in the r1Mems sequence, and 14 (the next after 18 in the r1Mems sequence) is actually remote from 18, as is neuron 5.  Of more interest are neurons 11 and 16, which are moderately distant in the sequence starting from 18.  Neuron 11 is five positions away from 18, and 16 is six positions away, in the r2Mems sequencing, and one less in each case in the sequencing from 18.

It is worth considering that in the original B-Matrix scheme, the activation sequences are handled purely according to the distance from the initial neuron, regardless of proximity of a neuron to a close neighbor.  For small models, this is likely more than sufficient as an assumption, as we must assume that the synapse latency and neuron response time are large compared to the speed of an electrical impulse through a dendrite.  But, we observe that for large models, involving nerve stimulation sequences that require multiple hops, this will not be a good assumption. For our present purposes we retain the assumption.  This avoids similar considerations with regard to neuron 19, which is somewhat remote from neuron 3, but the first stimulated by neuron 2.

The stimulation sequences, in relation to the respective r1Mems and r2Mems sequences, are as follows.

0

```
r1Mems = [ 1 0 0 0 0 0 0 0 0 -1 0 0 0 0 0 0 0 0 0]
r2Mems = [-1 0 0 0 0 0 0 0 0  0 0 0 0 1 0 0 0 0 0]
```

```
r1Mems = [ 1 1 0 0 0 0 0 1 0 0 -1 0 0 0 0 0 0 0 0 0]
r2Mems = [-1 1 0 0 0 0 0 0 0  0 1 0 0 1 0 0 0 0 0]
```

```
r1Mems = [ 1 1 1 0 0 0 0 1 0 1 -1 0 0 0 0 0 0 0 0 0]
r2Mems = [-1 1 1 0 0 1 0 0 0  0 1 0 0 1 0 0 0 0 0]
```

```
r1Mems = [ 1 1 1 1 0 1 0 1 0 1 -1 0 0 0 0 0 0 0 0 0]
r2Mems = [-1 1 1 1 0 1 0 0 0 0  0 1 0 0 1 0 1 0 0 0]
```

```
r1Mems = [ 1 1 1 1 1 1 0 1 0 1 -1 0 0 0 0 0 0 0 1 0]
r2Mems = [-1 1 1 1 1 1 0 0 0 0  0 1 0 0 1 0 1 1 0 0]
```

Notice that at this point in the r1Mems sequence the next neuron in the sequence is already clamped, and is a 1, not the expected and correct -1. This is, of course, due to the fact that neuron 13 was stimulated in iteration 3. Note also that this is the point at which the r2Mems stimulation sequence generates an erroneous state. At the same time, we have the same sequence element clamped in the r2Mems sequence, since neuron 17 was the third neuron stimulated in the r1Mems sequence, but is in this case correct in its state. This is not a surprise, since we already knew that the r1Mems sequence produced a correct third memory when stimulated in sequence. The net effect of this cross-clamping could be to skip an iteration, or to simply continue on with the next round affecting the states of the next sequential neurons. We choose the latter approach.

```
r1Mems = [ 1 1 1 1 1 1 -1 1 0 1 -1 0 0 0 0 0 0 0 1 0]
r2Mems = [-1 1 1 1 1 1 -1 0 0 0  0 1 0 0 1 0 1 1 0 0]
```

With the cross stimulation here, each sequence produces the same result in the same position, both correctly, for neuron 11. But, the next sequential stimulation in the r1Mems sequence is of an already clamped neuron (18), but no such condition in the r2Mems sequence. Therefore we pump the r2Mems sequence, but not the r1Mems sequence.

```
r1Mems = [ 1 1 1 1 1 1 -1  1 0 1 -1 -1 0 0 0 0 0 0 1 0]
r2Mems = [-1 1 1 1 1 1 -1 -1 0 0  0  1 0 0 1 0 1 1 0 0]
```

```
r1Mems = [ 1 1 1 1 1 1 -1  1  1 1 -1 -1 0 -1 0 0 0 0 1 0]
r2Mems = [-1 1 1 1 1 1 -1 -1 -1 0  0  1 0  0 1 1 1 1 0 0]
```

We again encounter a blocked condition in the r1Mems sequence, but with two consecutive positions. Therefore we pump the r2Mems sequence twice. But note here that even though we have been skipping rounds we will, after these two skips, still have the same number of positions not yet clamped. This is because all neurons exist in both sequences.

```
r1Mems = [ 1 1 1 1 1 1 -1  1  1 1 -1 -1 0 -1 0 0 1 0 1 0]
r2Mems = [-1 1 1 1 1 1 -1 -1 -1 1  0  1 0  0 1 1 1 1 0 0]
```

9

```
r1Mems = [ 1 1 1 1 1 1 -1  1  1 1 -1 -1 0 -1 0 0 1 -1 1 0]
r2Mems = [-1 1 1 1 1 1 -1 -1 -1 1 -1  1 0  0 1 1 1  1 0 0]
```

10

```
r1Mems = [ 1  1  1  1  1  1 -1  1   1  1 -1 -1 -1 -1 -1  0  1 -1   1  0]
r2Mems = [-1  1  1  1  1  1 -1 -1  -1  1 -1  1 -1   0  1  1  1  1 -1  0]
```

We skip a round in the r1Mems sequence again.

11

```
r1Mems = [ 1  1  1  1  1  1 -1  1   1  1 -1 -1 -1 -1 -1 -1  1 -1   1  0]
r2Mems = [-1  1  1  1  1  1 -1 -1  -1  1 -1  1 -1 -1   1  1  1  1 -1  0]
```

12

```
r1Mems = [ 1  1  1  1  1  1 -1  1   1  1 -1 -1 -1 -1 -1 -1  1 -1   1  1]
r2Mems = [-1  1  1  1  1  1 -1 -1  -1  1 -1  1 -1 -1   1  1  1  1 -1  1]
```

After 12 rounds, we have all the neurons clamped. Comparing the results with the memory expected (memory 3), we find both sequences have a single bit error for the same neuron, neuron 13. Concluding that this process is "faster" than a stimulus-chain from a single source would be incorrect, as will be discussed in the analysis that follows. At present, though, it should be sufficient to point out that the dendritic pathways from each of the initial neurons would remain active until their signals reached the various termini.

## Analysis

If we view the quantum events associated with the error that arose in neuron 13, we should note that the r1Mems sequence of stimulations would have attempted to drive neuron 13 to the appropriate status: -1. As a probabilistic event, we could have as easily performed a fair random draw to determine which stimulus event would have been determinative for that neuron. Since all other neurons reached the correct state, we would then have had a 0.5 probability of terminating with the correct memory. We might easily also include temporal effects, including an equilibration process at each neuron to deal with conflicts between stimuli. The "fair random draw" mentioned above is just one ultra-simple form of such a process.

Equilibration would continue throughout the network until all stimuli reached their connected termini, and those recipient neurons achieved equilibrium. Thus, a fuller examination of the subject broached here will include a much more detailed consideration of the various latencies that arise within and between neurons and their dendrites. This returns to the question raised with the stimulation of neuron 18 above. At what point does a stimulated neuron initiate its contribution to the stimulus cascade? This is compounded by a second question also alluded two, which is dealing with neural connectivity that is not "complete" in the sense that the number of dendrites associated with a neuron is much smaller than the number of neurons. Neural networks in real organisms are not simply direct-connect schemes, but involve layering, cross activation among large, multiply connected sub-graphs, etc. While these issues are of interest in the longer term, their examination would necessarily follow a great deal of other work in dealing with fully connected cases.

In terms of "performance", again, to conclude that the propagation of stimuli is in any sense completed when all neural states are clamped as in the example presented is a mistake. An equilibrium state would be reached only after all stimuli reached and were fully incorporated into their termini. However, from anecdotal experience, it is easy to draw an interesting parallel between the clamped state above and two experiences.  The first of these experiences is being reminded of something and taking a moment or more to clearly recall the full event or events, especially when the event(s) are complex and full of small details.  One might easily speculate that this is evidence of the equilibration process in action – and certainly should, at least as a hypothesis for further testing.

The second observation is that we know from personal experience that memory can be flawed, and that we can easily miss details when recalling events.  In environments where multiple stimuli trigger such recall, the example presented suggests that there may be a probabilistic element to the equilibration process that can result in one or more "bad" stimulus sequences corrupting "good" sequences, or being corrected by the "good" sequences.  This latter case also has anecdotal parallels in cases where two people recall and discuss the same event, and in the conversation remind each other of details missed or incorrectly recalled by the other.

In total, the question of how small subsets of neurons can trigger complex interplay between stimulus sequences to produce a "recall event" appears to be a field with rich potential, and should be examined in greater detail. This approach should also be compared with that of the instantaneously trained neural networks [10]-[12]. Paralleling spatial behavior of the B-matrices, certain feedforward networks will also display similar characteristics.